\documentclass{article}

\usepackage[preprint]{timeseries_workshop}
\usepackage{colortbl} 
\usepackage{booktabs} 
\usepackage{amsmath}

\usepackage[utf8]{inputenc} 
\usepackage[T1]{fontenc}    
\usepackage{hyperref}       
\usepackage{url}            
\usepackage{booktabs}       
\usepackage{amsfonts}       
\usepackage{nicefrac}       
\usepackage{microtype}      
\usepackage{xcolor}         
\usepackage{graphicx}
\usepackage{natbib}
\usepackage{subcaption}
\usepackage{wrapfig}
\usepackage{enumitem}

\title{Mamba4Cast: Efficient Zero-Shot Time Series Forecasting with State Space Models}

\author{%
  Sathya Kamesh Bhethanabhotla\thanks{Both authors contributed equally to this work.}  \\
  University of Freiburg\\
  \texttt{bhethans@tf.uni-freiburg.de} \\
  \And
  Omar Swelam\footnotemark[1]  \\
  University of Freiburg\\
  \texttt{swelamo@tf.uni-freiburg.de} \\
  \AND
  Julien Siems \\
  University of Freiburg\\
  \And
  David Salinas \\
  University of Freiburg\\
  \And
  Frank Hutter \\
  ELLIS Institute Tübingen \&\\ University of Freiburg\\
}

\begin{document}

\maketitle

\begin{abstract}
  This paper introduces Mamba4Cast, a zero-shot foundation model for time series forecasting. Based on the Mamba architecture and inspired by Prior-data Fitted Networks (PFNs), Mamba4Cast generalizes robustly across diverse time series tasks without the need for dataset specific fine-tuning. Mamba4Cast's key innovation lies in its ability to achieve strong zero-shot performance on \mbox{real-world} datasets while having much lower inference times than time series foundation models based on the transformer architecture. Trained solely on synthetic data, the model generates forecasts for entire horizons in a single pass, outpacing traditional auto-regressive approaches. Our experiments show that Mamba4Cast performs competitively against other state-of-the-art foundation models in various data sets while scaling significantly better with the prediction length.  The source code can be accessed at ~\url{https://github.com/automl/Mamba4Cast}.%
\end{abstract}

\section{Introduction}
    \label{submission}

    Time series forecasting is a critical task in numerous domains, from finance \citep{math11041054} to healthcare \citep{jung2021self}, and has been approached through various deep learning methods in recent years \citep{CHEN2023101819, liu2024deep}. Time-series data often exhibits complex temporal patterns, varying distributions with many confounding variables, and long-range dependencies, making it more challenging to model than other data paradigms. 
    Although the recent Cambrian explosion in deep learning, especially foundation models ~\citep{touvron2023llama, yu2022coca}, can be attributed in part to the availability of large amounts of data for training, the same cannot be said about forecasting in some domains \citep{NEURIPS2023_ae7d9c77, SIVAROOPAN2024110616}.
    
    Forecasting models \citep{DeepAR, zeng2022transformerseffectivetimeseries, oreshkin2019n} have traditionally employed non-zero-shot methods, which typically require customized training or fine-tuning for each specific task. While effective, this approach can be resource-intensive and time-consuming. Transformer-based time series foundation models \citep{ansari_chronos_2024, rasul_lag-llama_2024, dooley_forecastpfn_2023} have demonstrated significant potential to address these limitations. However, their application to long sequences during inference is constrained by their quadratic sample complexity. 
    
    In an effort to address both of these problems, we present Mamba4Cast, a time series foundation model based on two concepts: Prior-data Fitted Networks (PFNs)~\citep{hollmann_tabpfn_2023, dooley_forecastpfn_2023} and the Mamba~\citep{gu_mamba_2024, dao_transformers_2024} architecture. Our contributions are twofold:
    \begin{itemize}
        \item We introduce Mamba4Cast, a Mamba-based zero-shot forecasting model trained exclusively on synthetic data. It achieves 
        competitive performance compared to 
        other state-of-the-art zero-shot models, such as Chronos~\citep{ansari_chronos_2024}, while leveraging Mamba's architecture for efficient scaling over longer context lengths.
        \item We demonstrate that Mamba4Cast provides accurate point predictions over the entire forecast horizon in a single forward pass, achieving inference speeds several times faster than autoregressive counterparts.
    \end{itemize}

    \begin{figure}[t]
    \centering
    \vspace{-25pt}
    \includegraphics[trim={0.5cm 0.1cm 0.5cm 0.2cm}, clip, width=0.8\columnwidth]{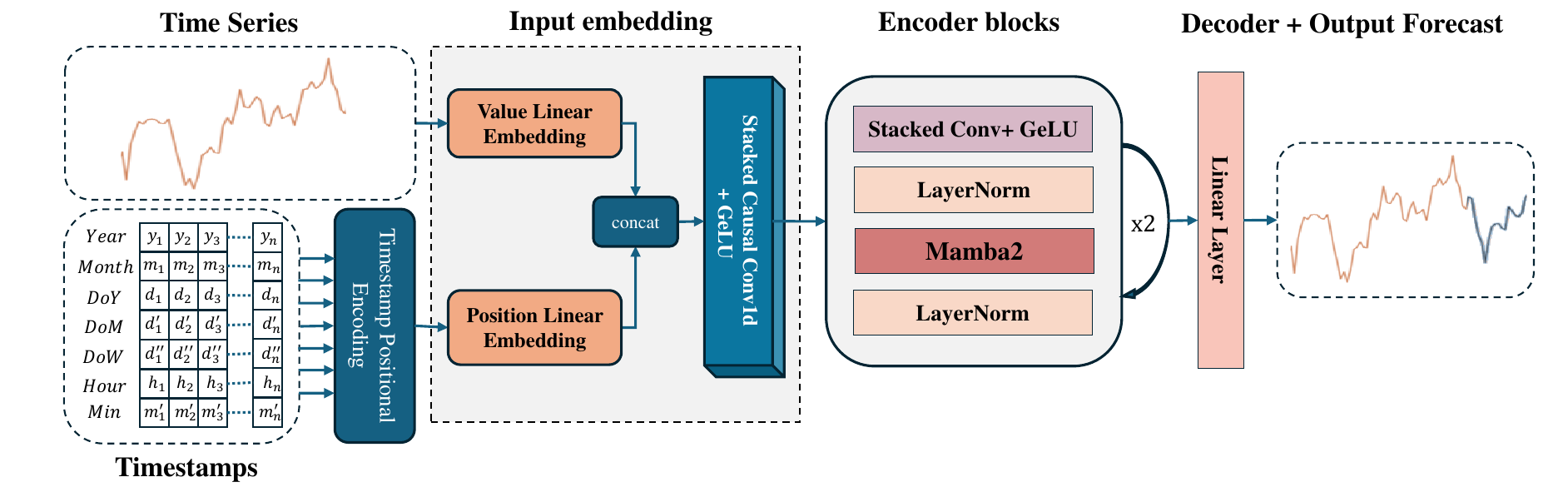}
    \caption{Schematic overview of the Mamba4Cast architecture.}
    \label{fig:mamba4cast_arch}
    \end{figure}

\section{Related Work}
    
    \textbf{Time series forecasting with Transformers} \enspace
    In the last few years, transformer-based models have significantly improved the state of the art in time series forecasting. Works like the Informer~\citep{zhou_informer_2021} and PatchTST~\citep{nie_time_2023} address the issue of long-term forecasting with transformers. \par

    \textbf{Zero-shot forecasting} \enspace
    There have also been several advancements in zero-shot time series forecasting  \citep{woo_unified_2024, gruver_large_2024}. %
    \citet{gao_units_2024} proposed UNITS, a unified multi-task model handling various predictive and generative tasks, and \citet{oreshkin_meta-learning_2020} proposed a meta-learning framework for zero-shot forecasting. These works highlight the growing trend towards more adaptable generalized time series models. \par

    \textbf{Forecasting LLMs} In the wake of the recent success of Large Language Models (LLMs), a novel direction in time series analysis has emerged, focusing on adapting LLM-based architectures for forecasting. Studies such as \cite{liu_autotimes_2024, jin_time-llm_2024, rasul_lag-llama_2024} have demonstrated the effectiveness of re-purposing LLMs for time series tasks. These approaches involve techniques to align time series data with the text-based input expected by LLMs, such as using text prototypes or encoding time series as strings of numerical digits. Notably, \cite{gruver_large_2024} showed that LLMs can perform zero-shot time series forecasting at levels comparable to or exceeding purpose-built models. These developments suggest that LLMs are promising candidates for general-purpose time series analysis, which can offer advantages in flexibility and performance in various forecasting tasks.\par
    
    \textbf{Training on Synthetic Data} \enspace While pre-training has enhanced the generalization capabilities of many models, their inductive biases often remain constrained to the distributions of their training corpus, potentially necessitating fine-tuning for niche applications. ForecastPFN \citep{dooley_forecastpfn_2023}, inspired by PFNs \citep{hollmann_tabpfn_2023, mullertransformers}, addressed this limitation by training on synthetic data, enabling zero-shot generalization to real-world time series. More recently, Chronos \citep{ansari_chronos_2024} demonstrated state-of-the-art results by training on both synthetic and real-world time series, introducing a transformer-based foundation model that follows the next-token prediction paradigm of large language models. \par
    
    \textbf{State Space Models} \enspace Despite the success of transformer-based methods, they face scalability challenges due to their quadratic complexity. In contrast, state-space models, such as Mamba \citep{gu_mamba_2024, dao_transformers_2024} or Linear Attention~\citep{katharopoulos2020transformers, yanggated, yang2024parallelizing}, have emerged as more efficient architectures, adapting state space models / linear RNNs~\citep{poppel2024xlstm} for sequence modeling with linear scaling properties. This efficiency has proven crucial for modeling dense, long-sequence data in vision and time series forecasting \citep{behrouz_mambamixer_2024, patro_simba_2024}. Subsequent works have further demonstrated Mamba's capacity in multivariate time-series forecasting; e.g., \citet{wang_is_2024} and \citet{liang_bi-mamba_2024}  proposed bi-directional Mamba architectures to capture inter- and intra-series dependencies, with the latter introducing a forget gate for enhancing selective performance on longer ranges. %
    With recent studies showcasing Mamba's in-context learning capabilities \citep{grazzi_is_2024, park_can_2024}, Mamba4Cast attempts to utilize them towards a Mamba-based foundation model for zero-shot time series forecasting. We aim to address this unexplored avenue for univariate time series, by training over a diverse set of synthetic generation procedures that generalize to various real-life datasets.

\section{Methodology}
    \subsection{Background: State Space Models}\label{subsec:state_space_models}
    
    Mamba4Cast builds upon the Mamba2 state-space model introduced by~\citet{dao_transformers_2024}. \mbox{Mamba2} is a linear Recurrent Neural Network described by the following recurrence:
    \begin{align*}
    h_t = A_t h_{t-1} + B_t x_t; \qquad
    y_t = C_t h_t
    \end{align*}
    where $h_t$, $x_t$, and $y_t$ represent the hidden state, input token embedding, and output at index $t$, respectively.
    In contrast to Mamba~\citep{gu_mamba_2024}, which uses a fully parameterized diagonal state transition matrix $A_t$, Mamba2 employs a scalar multiple of the identity matrix allowing for more efficient computation. The recurrence can be computed in chunks of linear attention blocks that can be pieced together later, leveraging tensor cores through matrix multiplication. This approach differs from Mamba's evaluation through an associative scan, which is also performed in parallel across the sequence but cannot leverage GPUs as well.
    
    \subsection{Mamba4Cast Architecture}
    Our proposed architecture, illustrated in Figure~\ref{fig:mamba4cast_arch}, consists of four primary components: \\(1) \textit{Pre-processing}: we scale the input series using a Min-Max Scaler and extract time features for positional embeddings. (2) \textit{Embedding}: we embed the scaled input values and their temporal information using convolutions with different dilations, ensuring a large receptive field for the representation used by future layers. For more details about data pre-processing and embedding, refer to Appendix~\ref{app:preprocessing}. (3) \textit{Encoder}: comprises of Mamba2 blocks with LayerNorm to avoid noisy learning signals followed by another dilated convolution layer.
    (4) \textit{Decoder}: the final component is a linear projection layer that transforms the embedded token representations into point forecasts.

    We perform an ablation study, detailed in Appendix~\ref{app.ablations}, investigating the role of convolutions, the efficacy of synthetic data generation methods, and the performance of alternative inference strategies.

    \subsection{Synthetic Data Generation} \label{sec.prior}
    The quality and diversity of the data generation process are crucial for Mamba4Cast's performance on real-world data, as it is trained exclusively on synthetic data. We employ two types of data-generating priors: ForecastPFN (FPFN) and Gaussian Process (GP) based.
    The \textit{FPFN prior}, based on \citet{dooley_forecastpfn_2023}, decomposes a time series into trend, seasonality, and noise components reflecting real-life patterns. 
    The \textit{GP prior}, inspired by \citet{ansari_chronos_2024}, complements the FPFN priors by accounting for patterns not captured therein. Each series is sampled from a GP with either a zero or a linear mean function and a composite kernel drawn from our \textit{Kernel bank}. This allows for generating diverse and realistic synthetic time series that exhibit a wide range of temporal behaviors. Detailed descriptions of these data priors are provided in Appendix ~\ref{app.synthetic}.

\section{Experiments}

    \begin{figure}[t]
        \centering
        \vspace{-35pt}
        \begin{subfigure}[b]{0.44\textwidth}
            \centering
            \includegraphics[trim={0.1cm 0.23cm 0.12cm 0.1cm}, clip,width=\textwidth, height=13em]{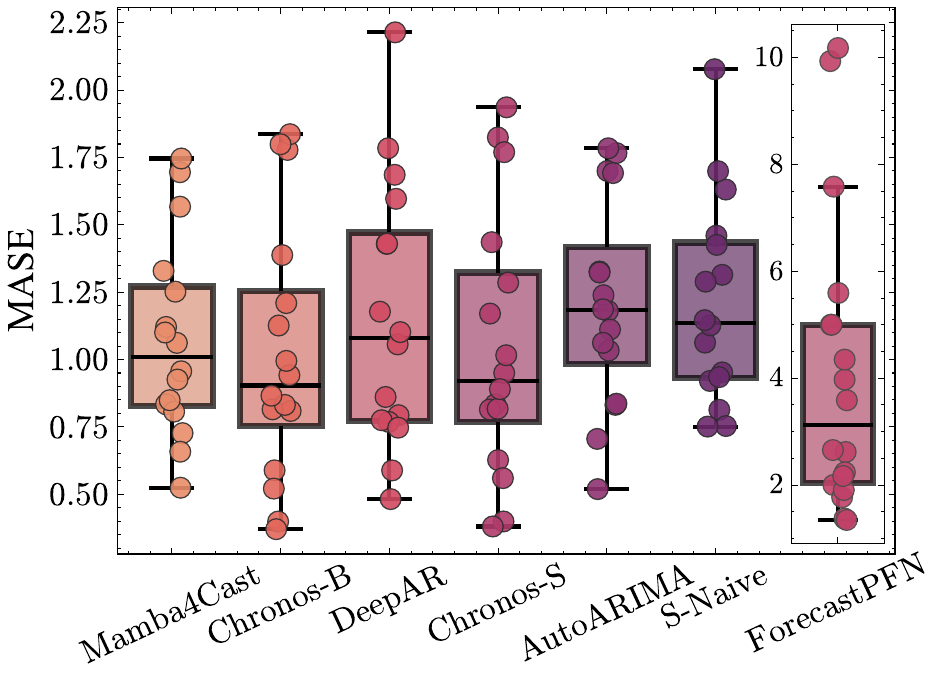}
            \label{fig:mase_box_plot}
        \end{subfigure}
        \hspace{2.5mm}
        \begin{subfigure}[b]{0.43\textwidth}
            \centering
            \includegraphics[trim={0.2cm 0.2cm 0.12cm 0.1cm}, clip, width=\textwidth]{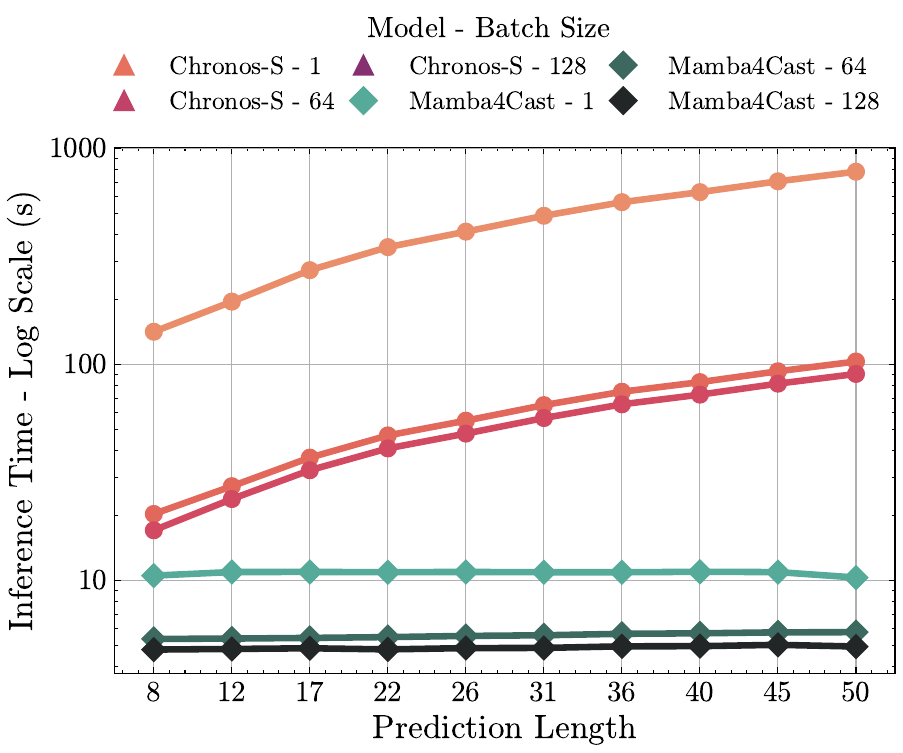}
            \label{fig:inf_time_plot}
        \end{subfigure}
        \caption{Performance and efficiency comparison of Mamba4Cast against baseline models. (\textit{left}) Distribution of MASE across 16 evaluation datasets (excluding Covid Deaths) for Mamba4Cast and five baseline models (ForecastPFN was much worse and is on a separate scale). (\textit{right}) Inference time of Mamba4Cast versus Chronos-Small on synthetically generated time series (2048 series, 512 context length) for increasing prediction lengths and varying batch sizes. The results demonstrate Mamba4Cast's superior efficiency, particularly for longer prediction horizons and larger batch sizes.}
        \label{fig:both_images}
        \vspace{-10pt}

    \end{figure}

    \subsection{Training Details} \label{sec.training}
    
    \paragraph{Architectural choices} The Mamba4Cast model is designed with approximately 27M parameters, positioning it between Chronos-Mini (20M) and Chronos-Small (46M) in size. As demonstrated in Figure \ref{fig:mamba4cast_arch}, Mamba4Cast is built on Mamba2 \citep{dao_transformers_2024} with a state expansion factor (N) of 128 and a block expansion factor (E) of 2. It features 2 encoder layers following an input projection to an embedding dimension of 1024. The final layer of the encoder is defined similarly to the stacked convolution layer illustrated in Appendix \ref{app:preprocessing} with the difference in the input channels being 1024 for the embedding size.  We minimize the mean squared error over the prediction horizon using AdamW \citep{loshchilov2019decoupledweightdecayregularization}. 
    \paragraph{Training setup} The model is trained for 420K batches of size 64, using data sampled from the priors in Section \ref{sec.prior}, via a parallelized data loader that ensures the same sample is not seen twice. We train on sequence lengths uniformly sampled between 30 and 512 and minimize the mean squared error over a prediction length uniformly sampled between 10 and 60 per batch. 50\% of the time we train to predict a contiguous chunk from the middle of the prediction length to improve predictability over the sequence by reducing reliance on previous states and encouraging emphasis on temporal information. The learning rate is cosine annealed \citep{loshchilov2017sgdr} from 1e-5 to 1e-7 throughout the training.

    The model is trained exclusively on synthetic data generated using two methods outlined in Section ~\ref{sec.prior}. The data composition is 70\% sampled from GP priors and 30\% sampled from FPFN priors, leveraging the GP kernels' flexibility in capturing diverse patterns. Training was conducted over 3 days on a single Nvidia RTX2080Ti GPU, for 360k training rounds consisting of 64 independently generated samples each. As stated in Appendix \ref{subapp:gp_prior}, we continue training for another 60K rounds with a changed kernel composition and a learning rate of 1e-6.

    \subsection{Performance Comparison with Baseline Models}\label{sec.baselines_comparisons}

        \begin{wrapfigure}{r}{0.5\textwidth}
        \vspace{-5pt}
        \includegraphics[trim={0cm 0.5cm 0cm 0.3cm}, clip, width=0.5\textwidth]{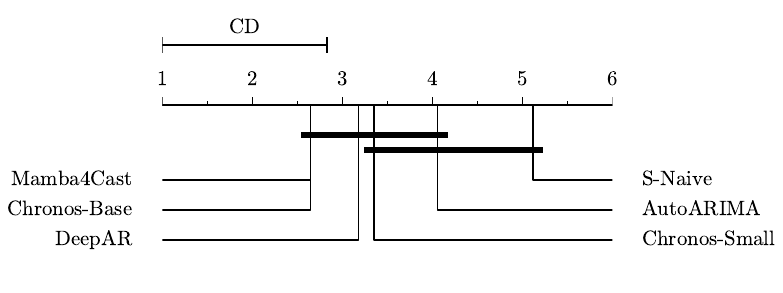}
        \caption{Critical difference diagram comparing mean MASE ranks of Mamba4Cast and baseline models across 17 time series datasets. ForecastPFN was much worse and is excluded for the sake of visibility. }
        \label{fig:crit_diff}
        \vspace{-10pt}
    \end{wrapfigure}
    
    We evaluate on 17 publicly available time series datasets from a wide range of domains from the dataset repository of the GluonTS \citep{alexandrov2019gluontsprobabilistictimeseries} library with a 512 context length. A detailed list of the datasets used is included in Appendix \ref{app.datasets}. Our evaluations involve comparisons with zero-shot baselines trained on synthetic data (Chronos and ForecastPFN), a deep learning baseline (DeepAR), and statistical methods (AutoARIMA and Seasonal Naive). 
        
    For our evaluations, we use AutoGluon–TimeSeries \citep{shchur2023autogluon} to evaluate the baselines, with the exception of ForecastPFN, whose results are sourced from the Chronos paper \citep{ansari_chronos_2024}. To ensure fair comparison across datasets with varying scales, we use the MASE metric \citep{masepaper}, which is scale-invariant.
    
    The results, as illustrated in Figures \ref{fig:both_images} and \ref{fig:crit_diff}, demonstrate that Mamba4Cast achieves competitive performance with Chronos-Base(200M) and surpasses other baselines. Notably, this performance is achieved without fine-tuning on real-world datasets. Figure \ref{fig:crit_diff} shows a critical difference diagram, visualizing the mean model rankings based on MASE (Mean Absolute Scaled Error) over the datasets. In this diagram, models are arranged from best (\textit{left}) to worst (\textit{right}), with statistically insignificant performance differences indicated by connecting horizontal lines (at a significance level of $\alpha = 0.05$). Detailed information on the MASE metric and per-dataset results can be found in Appendix~\ref{app.eval_real}.

    \subsection{Qualitative Analysis on Synthetic and Real Data}

\begin{figure}[t]
        \centering
        \begin{subfigure}[b]{0.48\textwidth}
            \centering
            \includegraphics[trim={0.1cm 0.2cm 0.08cm 0.1cm}, clip,width=\textwidth]{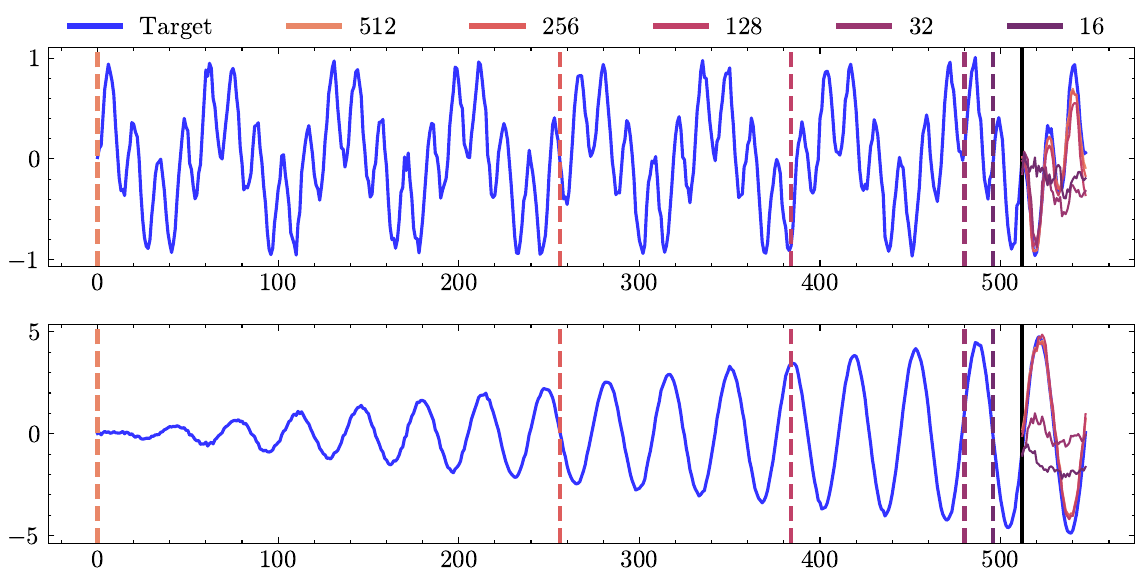}
            \label{fig:sine_exp_plot}
        \end{subfigure}
        \begin{subfigure}[b]{0.50\textwidth}
            \centering
            \includegraphics[trim={0.05cm 0.15cm 0.08cm 0.1cm}, clip,width=\textwidth]{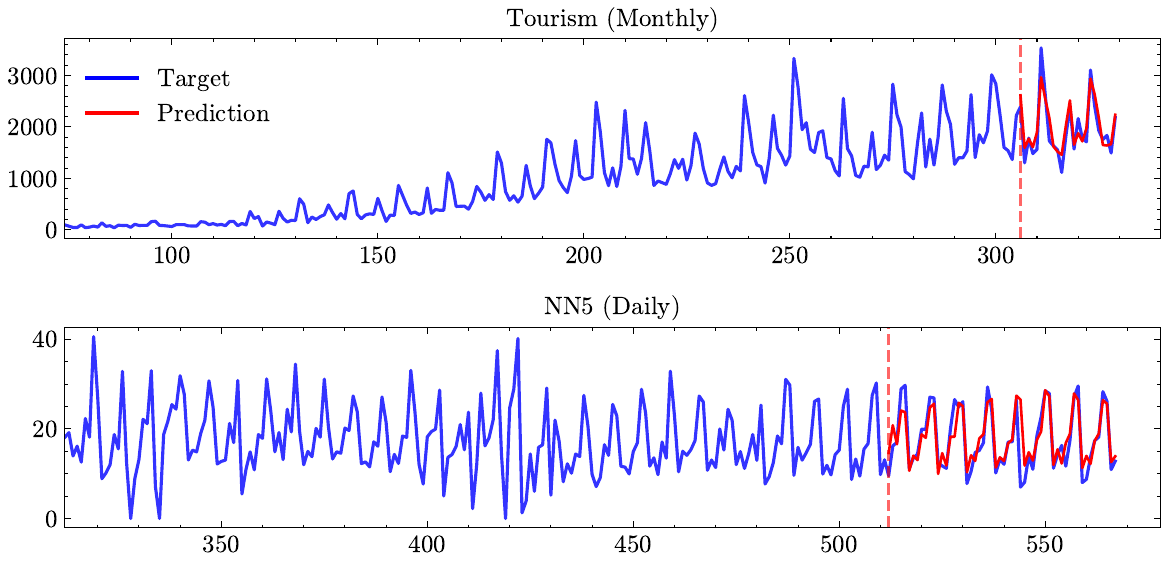}
            \label{fig:real_data_half_plot}
        \end{subfigure}
        \caption{Qualitative analysis of Mamba4Cast's performance. (\textit{left}) Demonstrates how prediction accuracy improves with increasing context length for multiplicative sine waves. (\textit{right}) Illustrates the model's forecasting capabilities on two real-world time series datasets.}
        \label{fig:prediction_plot}

    \end{figure}

    We conduct a qualitative inspection of Mamba4Cast to evaluate its ability to extrapolate over the forecasting horizon. Figure \ref{fig:prediction_plot} illustrates Mamba4Cast's improvement with increasing context length and its ability to capture real-life patterns. We also visualize the model's forecasting capability on additional real-world data in Appendix \ref{app.eval_real}.

\section{Conclusion and Future Work}

Our experiments demonstrate Mamba's capability in creating a reliable zero-shot time-series foundation model. After training solely on synthetic data, Mamba4Cast achieves near state-of-the-art results while also maintaining scalability and efficient inference. However, Mamba4Cast is limited to the univariate domain, which only forms a small portion of real time series problems, and is heavily reliant on the diversity of its priors. Nevertheless, we believe our work serves as a significant step towards developing highly performant and scalable multivariate zero-shot forecasting models, setting the stage for future advancements in this domain.

\section*{Acknowledgments}
This research was partially supported by the following sources: TAILOR, a project funded by EU Horizon 2020 research and innovation programme under GA No 952215; the Deutsche Forschungsgemeinschaft (DFG, German Research Foundation) under grant number 417962828; the European Research Council (ERC) Consolidator Grant “Deep Learning 2.0” (grant no. 101045765). The authors acknowledge support by the state of Baden-Württemberg through bwHPC and the German Research Foundation (DFG) through grant INST 35/1597-1 FUGG.Frank Hutter acknowledges financial support by the Hector Foundation. The authors acknowledge support from ELLIS and ELIZA. Funded by the European Union. Views and opinions expressed are however those of the author(s) only and do not necessarily reflect those of the European Union or the ERC. Neither the European Union nor the ERC can be held responsible for them.
\begin{center}\includegraphics[width=0.3\textwidth]{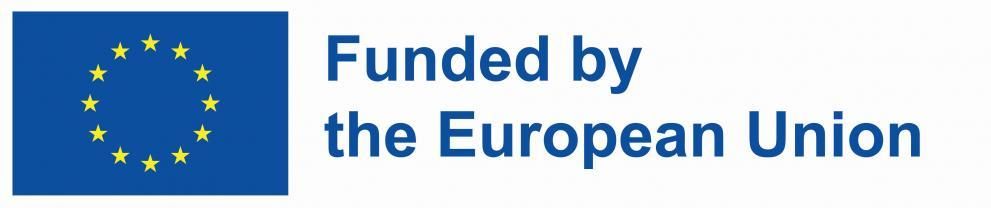}\end{center}. 

\bibliographystyle{icml2024}
\bibliography{example_paper}

\newpage
\appendix

\section{Synthetic Data Generation}\label{app.synthetic}

    \subsection{ForecastPFN Prior}
        We adopted the prior generation process from \citet{dooley_forecastpfn_2023} that decomposes the time series into three components as outlined in Section \ref{sec.prior}. The trend incorporates linear and exponential growth factors, while seasonal components capture periodic variations at multiple time scales (minutely, hourly, daily, weekly, and monthly), reflecting natural cycles in the data. Noise is modeled using a Weibull distribution to maintain a constant expected value. We introduced some modifications to the original procedure that are mentioned below.
        
        \paragraph{Trend} In our experiments,  we found that training Mamba4Cast on long sequence time series with exponential trends results in suboptimal performance. Therefore we limited the non-linear growth behavior to be polynomial ones represented in the GP priors, while the FPFN prior only models linearly growing signals.
    
        \paragraph{Seasonal} Seasonal patterns are generated according to the granularity of the timestamps. For each granularity, we sample sine-wave signals, referred to as harmonics, with periodicities corresponding to that granularity: 60 for minutely, 24 for hourly, 7 for daily, and 12 for monthly data. For each time series, we sample harmonics from both its granularity and the immediate higher granularity. As an example, for minutely data, we sample seasonal signals with both minutely and hourly periodicities.
        In the original design, 10 or 6 harmonics were sampled for each granularity, but in our optimal setup, we used 8 and 5 harmonics, respectively. As the number of harmonics increases, their periodicity is scaled down by the harmonic index, allowing the model to capture finer fluctuations in the data.
    
    \subsection{GP priors} \label{subapp:gp_prior}

    \paragraph{GP model} We use GPyTorch \citep{gardner2018gpytorch} for sampling our time series from composite kernels, with a sampled zero or linear mean, and a Gaussian likelihood. We add the noise using Cholesky jitter, with the jitter level being sampled among 0.1, 0.01, and 0.001, with probabilities of 0.1, 0.2, and 0.7 respectively. This design choice is to generalize Mamba4Cast for different noise levels in the real-life datasets. 
    
    \paragraph{Kernels}  The kernel bank comprises Linear, Polynomial, Matern, and Periodic kernels. To ensure complex time series patterns, we combine up to 6 kernels using sampled \textit{binary operations} (addition or multiplication). The best training pipeline involved sampling a number of kernels from 1 to 6. In the first 360K training rounds, for each kernel, we sampled from Periodic, Matern, or Linear kernels with weights of 5, 1.5, and 1 respectively. This prior was inspired by the KernelSynth method outlined by Chronos \citep{ansari_chronos_2024}.\\
    We also observed that training followed by a \textit{fine-tuning} phase of 60K rounds with changed weights of Periodic, Matern, and Polynomial kernels to 5, 2 and 1 respectively, resulted in better generalization.

    \subsection{Signal level noises}
    In addition to the white noise signal incorporated in our priors, we introduce two types of \mbox{multiplicative} noise signals: spikes and step noise. \textbf{Spikes} are designed to introduce regular peaks at every interval, \textit{l}. To simulate peaks that occur regularly but are irregularly spaced, we apply a masking window \textit{m}, which masks up to 40\% of the spikes within the window. Similarly, multiplicative \textbf{step} functions are applied in an alternating high-low-high-low pattern to enable Mamba4Cast to capture seasonal level shifts.

    \section{Dataset Preprocessing}\label{app:preprocessing}
   
    We adopt a preprocessing approach similar to e.g. ForecastPFN~\citep{dooley_forecastpfn_2023}. Time-steps are decomposed into minutely, hourly, day of week, day of month, day of year, monthly and yearly components, encoded using sinusoidal embeddings. These encodings, along with the series value, are linearly projected and concatenated to represent each time-point in a 112 embedding vector. The value of target tokens for model input across the prediction horizon is 0 for the prediction of point value or 1 for the cumulative mean prediction, fixed to 0 during inference.
    
    With all input and output token embeddings stacked along the sequence dimension, we apply four causal convolution layers with kernel sizes of 5 and dilations of 1, 2, 4, and 8, concatenating their outputs for diverse temporal coverage. This facilitates capturing multi-scale temporal dependencies, enhancing our model's forecasting capabilities. The stack of causal convolution projects the tokens up into our desired embedding dimension of 1024 followed by an inception layer to combine the information across the temporal multi-scale for each token while maintaining the embedding size.

\section{Benchmark Datasets}\label{app.datasets}

    We use 17 datasets from Chronos zero-shot benchmark while removing datasets with very small context and prediction length, datasets that are very large, and datasets with sub-hourly frequencies. We will extend to support those datasets in future work. We used GluonTS as an interface for these datasets to have a comparable evaluation pipeline to Chronos. The context length (input sequence length) was restricted to be at most 512, while the prediction length varied according to the evaluated dataset as shown in Table \ref{table.evaluation_datasets}.   

    \begin{table}[t]
    \caption{Characteristics of Datasets Used for Zero-Shot Evaluation of Mamba4Cast and baselines.}
    \label{table.evaluation_datasets}
    \vskip 0.15in
    \begin{center}
    \begin{small}
    \begin{tabular}{lccc}
    \specialrule{1.2pt}{0pt}{0pt}
    \rule{0pt}{1.5em} \raisebox{0.3em}{\textbf{Dataset}} & \raisebox{0.3em}{\textbf{Frequency}} & \raisebox{0.3em}{\textbf{Num. Test Series}}& \raisebox{0.3em}{\textbf{Prediction Length}}\\
    \specialrule{1.2pt}{0pt}{0pt}
    \addlinespace[0.3em]
    CIF 2016        & 1M & 72 & 12\\
    Car Parts       & 1M & 2674 &12 \\
    Covid Deaths     & 1D & 266 & 30\\
    ERCOT Load      & 1H & 8 & 24\\
    Exchange Rate   & 1B & 8 & 30\\
    FRED-MD         & 1M & 107 & 12\\
    Hospital        & 1M & 767 & 12\\
    M1 (Monthly)    & 1M & 617 & 18\\
    M1 (Quarterly)  & 3M & 203 & 8\\
    M3 (Monthly)    & 1M & 1428 & 18\\
    M3 (Quarterly)  & 3M & 756 & 8\\
    NN5 (Daily)     & 1D & 111 & 56 \\
    NN5 (Weekly)    & 1W & 111 & 8\\
    Tourism (Monthly)   & 1M & 366 & 24\\
    Tourism (Quarterly) & 3M & 427 & 8\\
    Traffic         & 1H & 862 & 24\\
    Weather         & 1D & 3010 & 30 \\
    \bottomrule
    \end{tabular}
    \end{small}
    \end{center}
    \vskip -0.1in
    \end{table}

\section{Ablation studies} \label{app.ablations}

We investigate the robustness of Mamba4Cast in different configurations, which fall into three main categories:

    \begin{itemize}
        \item \textbf{Architectural Changes:} We look into the effectiveness of a stacked causal convolutions layer (CNN) against a linear layer (Linear) in the input embedding and as the encoder-block's last layer.  While adding the CNN layer as the final layer of the encoder block (\textit{baseline}) provides superior performance with a significant overhead in model size, the key advantage stems from the CNN layer in the input embedding without overhead in model size.
        
        The model sizes of the three setups listed in the corresponding section of Table \ref{tab.ablation_study} are 27M, 17M, and 15M, in the same order as in the table.
        
        \item \textbf{Prior Mixing Ratios:} Given the importance of the distribution of synthetic data, we conducted experiments to explore the impact of each of the two approaches mentioned in Section \ref{sec.prior}. The ablation indicates the effectiveness of the GP prior over the FPFN prior, leading to our choice of a GP favoured mixture of data for training.
        
        \item \textbf{Inference Modes:} Mamba4Cast was designed with efficient zero-shot forecasting in mind following the one-pass multipoint setup, in which the input and target tokens are concatenated together in their respective order.  Mamba4Cast also supports autoregressive forecasting, but its performance declines significantly in this setup. A likely reason is that feeding predicted values back into the model causes overconfidence and error propagation. In contrast, the multipoint setup treats all target values as unknown, avoiding this issue. 
        
        We further test the impact of ensembling by averaging the forecasts generated at 5 different levels of dropout, from 0 to 0.5, of the input sequence. However, given the superior performance over longer and more inclusive contexts, demonstrated in Figure \ref{fig:prediction_plot}, it follows that including a less accurate forecast can degrade performance in case Mamba4Cast is certain about its forecast as shown in Table \ref{tab.ablation_study}.
        
        \end{itemize}

        The ablation studies were conducted on the first 360K training rounds mentioned in Section \ref{sec.training}, as the subsequent 60K were later applied to our chosen setup for the baseline comparisons cited in Section \ref{sec.baselines_comparisons}.

\renewcommand{\arraystretch}{1.1} %

\begin{table}[h!]
\centering
\captionsetup{width=0.7\textwidth}
\caption{Ablation study on architectural changes, prior mixing ratios and the inference modes. The value reported is the geometric mean of MASE across all 17 datasets for each setup.}
\label{tab.ablation_study}
\vskip 0.15in
    \resizebox{0.6\textwidth}{!}{
    \begin{tabular}{lc}
    \specialrule{1.2pt}{0pt}{0pt}
    \rule{0pt}{1.5em} \raisebox{0.3em}{\textbf{Ablation Setup}} &  \raisebox{0.3em}{\textbf{Mean MASE}} \\ 
    
        \specialrule{1.2pt}{0pt}{0pt} 
        \addlinespace[0.3em]
        \raisebox{0.1em}{\textbf{Architectural Modifications}} & \rule{0pt}{1em} \\
        \cmidrule(r){1-2}
        \addlinespace[0.3em]
        $\text{CNN}_{in\_emb}$ / $\text{CNN}_{enc\_out}$ (\textit{Baseline}) & $1.153$ \\ 
        $\text{CNN}_{in\_emb}$ / $\text{Linear}_{enc\_out}$ & $1.205$ \\ 
        $\text{Linear}_{in\_emb}$ / $\text{Linear}_{enc\_out}$ & $1.556$ \\ 
        \addlinespace[0.3em] %
        
        \cmidrule(r){1-2}
        \raisebox{0.1em}{\textbf{Priors Mixing Ratios}} & \rule{0pt}{1em} \\ 
        \cmidrule(r){1-2}
        \addlinespace[0.3em]
        \makebox[2em][r]{70\%} GP Prior / \makebox[2.2em][r]{30\%} FPFN Prior (\textit{Baseline}) & $1.153$ \\
        \makebox[2em][r]{100\%} GP Prior / \makebox[2.2em][r]{0\%} FPFN Prior & $1.167$ \\
        \makebox[2em][r]{0\%} GP Prior / \makebox[2.2em][r]{100\%} FPFN Prior & $1.579$ \\ 
        \addlinespace[0.3em] %
        
        \cmidrule(r){1-2}
        \raisebox{0.1em}{\textbf{Inference Modes}} & \rule{0pt}{1em} \\ 
        \cmidrule(r){1-2}
        \addlinespace[0.3em]
        Multipoint Forecasting (\textit{Baseline}) & $1.153$ \\
        Autoregressive Forecasting & $2.044$ \\
        Ensemble Forecasting & $1.558$ \\ 
        \addlinespace[0.3em]
    \specialrule{1.2pt}{0pt}{0pt} 
    \end{tabular}}
\end{table}

\section{Evaluations on real datasets}\label{app.eval_real}

\subsection{Evaluation metric} As part of our evaluation, we tested the performance of our model on real-world time series datasets alongside the synthetic data. The primary metric used was the seasonal Mean Absolute Scaled Error (MASE), which scales the forecast error by the mean absolute error of a seasonal naïve forecast on the training data. The evaluation of Mamba4Cast on real-world datasets demonstrates the model's capability to generalize and perform well in diverse, real-world forecasting scenarios. Detailed evaluations per dataset can be found in Table \ref{tab.mase-per-dataset}. We witnessed inconsistencies between the evaluations performed by AutoGluon in our setups and the ones reported in Chronos paper on datasets with daily frequency, specifically on "Covid Deaths." This resulted in the large gap witnessed on ForecastPFN's results reported here, since the model's MASE evaluations are sourced from the Chronos paper. The results reported for Mamba4Cast per dataset are evaluated with the best model trained according to the procedures in Section \ref{sec.training}.

\begin{table}[h!]
\centering
\captionsetup{width=\textwidth}
\caption{MASE evaluations on all of the 17 datasets with the lower value the better. The best results per dataset are in bold and the second best results are underlined.}
\vskip 0.15in

\label{tab.mase-per-dataset}
    \resizebox{1\textwidth}{!}{\begin{tabular}{lcccccccc}
    \specialrule{1.2pt}{0pt}{0pt}
    \rule{0pt}{1.5em} 
    & \multicolumn{4}{c}{\textbf{Zero-shot}} 
    & \multicolumn{2}{c}{\textbf{Task-specific}} 
    & \textbf{Statistical Baseline} \\
    \cmidrule(lr){2-5} \cmidrule(lr){6-7} \cmidrule(lr){8-8}
    
    \rule{0pt}{1.5em} & \raisebox{0.3em}{\textbf{Mamba4Cast}} & \raisebox{0.3em}{\textbf{Chronos-B}} &  \raisebox{0.3em}{\textbf{Chronos-S}} &  \raisebox{0.3em}{\textbf{ForecastPFN}} &  \raisebox{0.3em}{\textbf{DeepAR}} & \raisebox{0.3em}{\textbf{AutoARIMA}} & \raisebox{0.3em}{\textbf{S-Naive}}\\
    \raisebox{0.3em}{\textbf{Dataset}} &  &  &  &  &  &  & \\
    \specialrule{1pt}{0pt}{0pt}
    Car Parts & 1.061 & 0.832 & \underline{0.817} & 2.657 & \textbf{0.747} & 1.180 & 1.127\\
    CIF 2016 & \textbf{0.925} & \underline{0.995} & 1.016 & 3.558 & 1.597 & 1.062 & 1.289\\
    Covid Deaths & \textbf{5.926} & 7.461 & 7.376 & 91.515 & 8.917 & \underline{6.059} & 8.977\\
    ERCOT Load & 0.657 & \textbf{0.521} & \underline{0.560} & 3.975 & 1.429 & 1.112 & 0.751\\
    Exchange Rate & \underline{1.329} & 1.388 & 1.436 & 7.583 & 2.214 & \textbf{1.187} & 1.460\\
    FRED-MD & 0.524 & \textbf{0.399} & \underline{0.399} & 2.621 & 0.588 & 0.519 & 0.935\\
    Hospital & \underline{0.806} & 0.815 & 0.814 & 1.775 & \textbf{0.775} & 0.836 & 0.921\\
    M1 (Monthly) & \textbf{1.100} & 1.126 & 1.171 & 2.172 & \underline{1.102} & 1.239 & 1.314\\
    M1 (Quarterly) & \textbf{1.695} & 1.778 & 1.824 & 9.931 & 1.784 & \underline{1.766} & 2.078\\
    M3 (Monthly) & \textbf{0.849} & \underline{0.866} & 0.890 & 2.240 & 1.056 & 1.033 & 1.146\\
    M3 (Quarterly) & 1.251 & \underline{1.210} & 1.285 & 10.176 & \textbf{1.178} & 1.323 & 1.425\\
    NN5 (Daily) & 0.833 & \underline{0.809} & 0.834 & 1.375 & \textbf{0.793} & 0.832 & 0.952\\
    NN5 (Weekly) & 0.956 & \underline{0.942} & 0.950 & 1.349 & \textbf{0.861} & 1.700 & 1.063\\
    Tourism (Monthly) & \underline{1.567} & 1.836 & 1.936 & 4.348 & \textbf{1.430} & 1.692 & 1.631\\
    Tourism (Quarterly) & 1.746 & 1.799 & 1.770 & 5.595 & \textbf{1.686} & 1.784 & \underline{1.699}\\
    Traffic & 1.120 & \textbf{0.370} & \underline{0.380} & 1.909 & 0.482 & 1.327 & 0.753\\
    Weather & 0.726 & \textbf{0.589} & \underline{0.627} & 2.003 & 0.769 & 0.705 & 0.813\\
    \specialrule{1.2pt}{0pt}{0pt}
    \end{tabular}}
\end{table}

\subsection{Qualitative analysis}\label{app:subsec:qualitative_analysis}
\begin{figure}[ht]
        \centering
        \includegraphics[width=0.98\columnwidth]{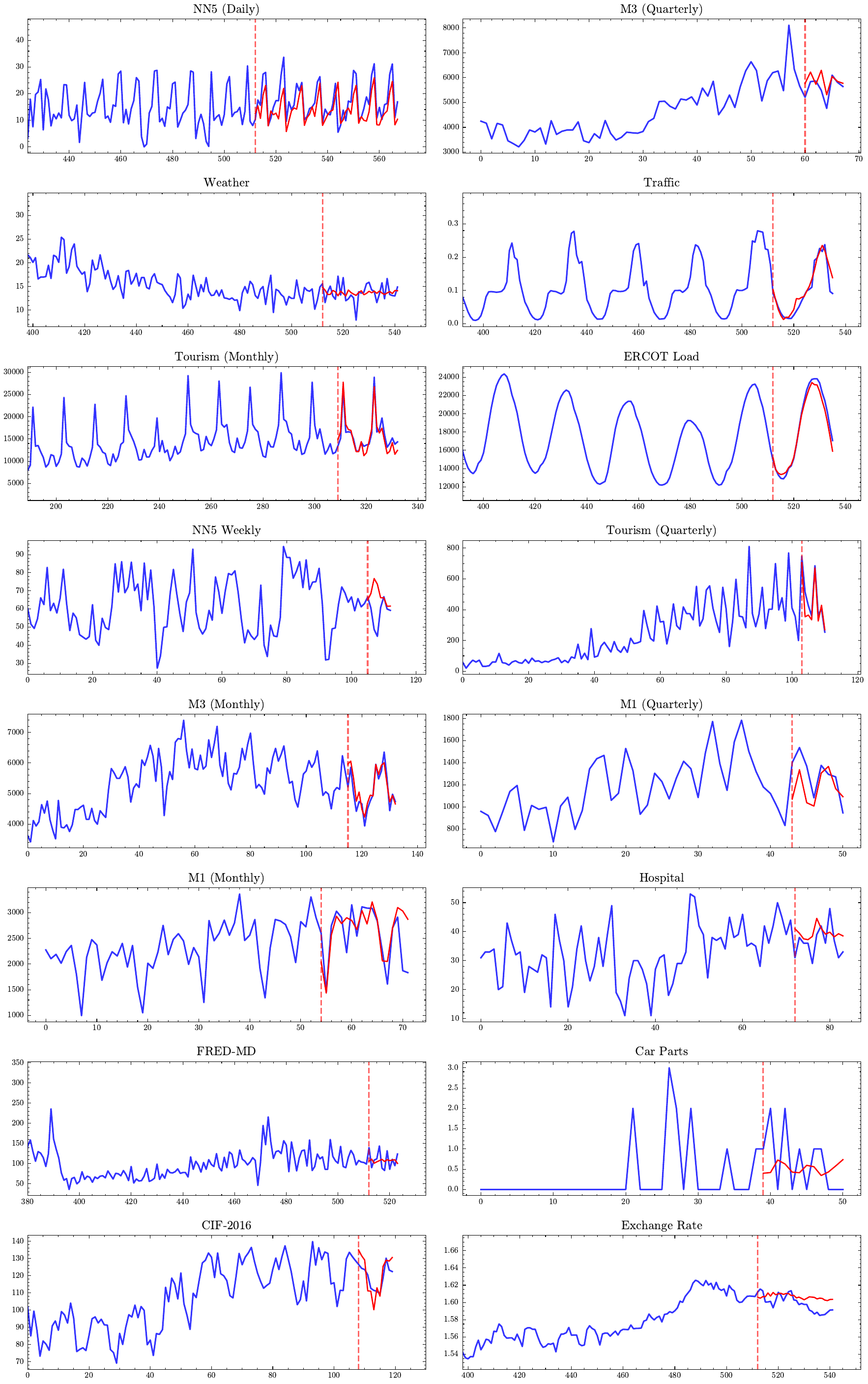}
        \caption{Qualitative analysis of real-world datasets evaluated by Mamba4Cast. Blue denotes the ground-truth, red the prediction.}
        \label{fig:real_preds}
\end{figure}

An impartial evaluation in time series forecasting applications favors a qualitative evaluation over the datasets in question, to guarantee adequate behavior for point forecasting. For this sake, Figure \ref{fig:real_preds}  demonstrates Mamba4Cast's ability to capture diverse patterns exemplified in the real-life datasets.
    
\end{document}